# Combining shape and contour features to improve tool wear monitoring in milling processes

María Teresa García-Ordás*, Enrique Alegre Gutiérrez, Víctor González-Castro and Rocío Alaiz-Rodríguez

*Department of Electrics, Systems and Automatics Engineering. Universidad de León. Campus de Vegazana s/n, 24071, León (Spain)*

In this paper, a new system based on combinations of a shape descriptor and a contour descriptor has been proposed for classifying inserts in milling processes according to their wear level following a computer vision based approach. To describe the wear region shape we have proposed a new descriptor called ShapeFeat and its contour has been characterized using the method BORCHIZ that, to the best of our knowledge, achieves the best performance for tool wear monitoring following a computer vision-based approach. Results show that the combination of BORCHIZ with ShapeFeat using a late fusion method improves the classification performance significantly, obtaining an accuracy of 91.44% in the binary classification (i.e., the classification of the wear as high or low) and 82.90% using three target classes (i.e., classification of the wear as high, medium or low). These results outperform the ones obtained by both descriptors used on their own, which achieve accuracies of 88.70% and 80.67% for two and three classes, respectively, using ShapeFeat and 87.06% and 80.24% with B-ORCHIZ. This study yielded encouraging results for the manufacturing community in order to classify automatically the inserts in terms of their wear for milling processes.

## 1. Introduction

Tool wear is one of the most influential factors on the quality of machined surfaces. Thus, a key step of a machine system is the replacement of tools affected by wear at the optimal moment. It is essential, not only because of the cost of cutting tools themselves, but also for the indirect costs derived by the fact that the machine must be stopped during the time the tool is replaced. For this reason, tool wear monitoring becomes a critical operation in automatic manufacturing. Many research groups have been working with the aim of developing automatic systems to deal with this problem (J. et al. 2015; Sick 2002; Jeang and Yang 1992; Zhou, Chandra, and Wysk 1990).

Approaches to tool wear monitoring may be divided into two broad categories: indirect and direct methods. Indirect monitoring methods estimate the wear by measuring variables such as cutting forces (Azmi 2015; Wang et al. 2014a), vibration (Li 2002) (Rao, Murthy, and Rao 2014) or acoustic emission (Scheffer and Heyns 2001). For example, Kilundu et al. use pseudo-local Singular Spectrum Analysis (SSA) on vibration signals measured on the tool holder in order to determine the tool wear (Kilundu, Dehombreux, and Chiementin 2011). This is coupled to a band-pass filter to allow definition and extraction of features which are sensitive to tool wear. These features are defined, in some frequency bands, from sums of Fourier coefficients of reconstructed and residual signals obtained by SSA. More recently, Nouri et al. (Nouri et al. 2015), developed a new method to monitor end milling tool wear in real-time by tracking force model coefficients during the cutting process. The behavior of these coefficients is shown to be independent from the cutting conditions and correlated with the wear state of the cutting tool. A multiple sensor monitoring system comprising cutting force, acoustic emission and vibration sensing units was employed in (Segreto, Simeone, and Teti 2013) for tool state assessment during turning of Inconel 718 nickel alloy. Feature extraction was carried out by processing the detected sensor signals in order to reduce the high dimensionality of the data. The extracted features were merged by means of a sensor fusion methodology based on neural network pattern recognition for decision making on tool wear condition.

The downside of indirect methods are that they do not provide sufficient information to perform an optimal replacement of the inserts because the relationship between tool wear and the observed variables depends on the cutting conditions and, in general, it is not known in advance. Even though these methods are the most popular, the precision achieved with them is seriously affected by noise signals in industrial environments. Furthermore, they are cost-inefficient and may be unavailable for some real applications (Kassim, Mannan, and Zhu 2007).

In contrast, direct methods based on computer vision monitor the state of the cutting tools directly at the cutting edge when the head tool is in a resting position. Although less popular than indirect approaches, they have the advantage of measuring actual geometric changes in the tool, offering more accuracy and reliability (Sick 2002) (Kurada and Bradley 1997).

In (Castejon et al. 2007), a binary image for each of the wear flank images have been obtained by applying several pre-processing and segmenting operations. Every wear flank region has been described by means of nine geometrical descriptors. LDA (Linear Discriminant Analysis) shows that three out of the nine descriptors provide the 98.63% of the necessary information to carry out the classification, which are eccentricity, extent and solidity. A finite mixture model approach shows the presence of three clusters using these descriptors, which correspond with low, medium and high wear level. A monitoring approach is performed using the tool wear evolution for each insert along machining and the discriminant analysis. Another work in which computer vision is employed for tool wear monitoring is the one developed by Chethan et al. (Chethan et al. 2014) in which a methodology to calculate the tool area based on the drill image thresholding is presented. This helps to select optimal drilling parameters in relation to tool wear. Datta et al. proposed a method based on texture analysis and Voronoi tessellation in order to measure progressive tool wear (Datta et al. 2013). Another method based on textures is explained in (Kassim, Mannan, and Zhu 2007). Some works attempt to describe the wear taking into account the wear contour (García-Ordás et al. 2014; García-Ordás et al. 2016). Both methods are based on the ZMEG contour descriptor (Anuar, Setchi, and kun Lai 2013) obtaining promising results in the tool wear monitoring field. B-ORCHIZ is a contour descriptor which combines global and local description.

Changes on the shape of a cutting tool depend on different types of wear (Zhang et al. 2011). A cutting tool experiences a very complex pattern of wear as it cuts the material and some parts of the cutting tool can loose material due to wear mechanisms. The wear shape is also different depending on the machining process type. For example, in milling processes the wear shape is distributed uniformly along the insert while in lathe processes the wear occurs over the same area concavely.

In this study, we explore to merge features that describe contour and features with shape information to yield a more powerful description. Both descriptors provide very useful information about the images but each one focuses on different characteristics. The possibility of using both to determine the the tool wear offers a new chance of improvement in this field. In particular, a new shape descriptor called ShapeFeat, that characterizes the wear based on the region properties is presented. It is also combined with the best contour descriptor found in the literature with this same purpose using three different fusion techniques.

This proposal has been assessed using images acquired in a laboratory under controlled conditions but it can be easily implemented in a real industrial environment. Fig. 1 shows the milling machine used in this study, the TECOI TRF milling machine and a close-up of the milling head tool, which is used to manufacture metal poles of wind towers. The monitoring process of the milling head tool is carried out during the period of time when it is parked, which lasts for at least five minutes, sometimes up to 20 minutes, before the processing of the next metallic plate. These machines use an aggressive edge milling in a single pass for the machining of thick plates, what makes tool wear appear even with low machining time. The acquisition would be carried out with a fixed camera and an illumination system composed by two led bars. A cover will enclose the system in order to attenuate the external illumination and, thus, to avoid light changes in the acquired images. A diagram depicting the whole acquisition system is shown in Figure 2. The location of the inserts may be obtained using the method explained in (Fernández-Robles et al. 2017). The whole process takes less than 1.5 minutes on a personal computer with a 2GHz processor and 8GB RAM. That includes the time since the image acquisition takes place until the cutting edges of the inserts are classified. Therefore, the inspection does not require to stop the machining longer than its normal resting time and it provides information about the state of each insert. In this way, the operator can replace the worn inserts and, therefore, make sure that the next metallic plate is processed under optimal conditions.

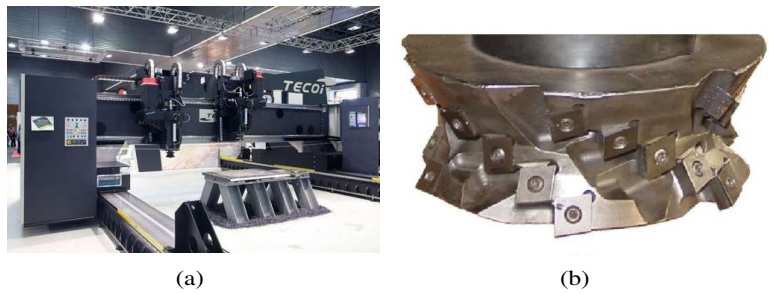

(a)        (b)

Figure 1. (a) Overview of the TECOI TRF milling machine and (b) detail of the milling head tool that we used in the experiments carried out in this work.

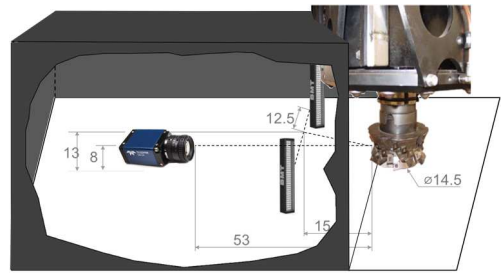

Figure 2. Diagram depicting an example of an acquisition system in a real environment.

The rest of the paper is organized as follows. In Section 2, the wear description methodology based on shape and contour features is presented. The combinational methods (early, intermediate and late fusion strategies) are introduced in Section 3. In Section 4, the dataset creation is described and the experimental setup and results for contour, shape and both features are discussed. Finally, we present the conclusions of our work in Section 5.

## 2. Methodology. Wear description with contour and shape features.

In this section, two image description methods are presented. The first one, ShapeFeat, is a new shape descriptor proposed in this work. The other one, B-ORCHIZ (García-Ordás et al. 2016), is a contour descriptor that provides the best performance in the literature up to this moment.

### 2.1 Shape Descriptor: ShapeFeat

In this paper, we propose a shape descriptor called ShapeFeat that takes into account ten different shape features extracted from the binary region of an image. This descriptor provides useful information about the image which is not usually obtained with descriptors based on moments, such as Hu (Hu 1962), Flusser (Flusser 1992) or Zernike (Zernike 1934), which are focused on the shape information, but avoid relevant information like the area of the region or its homogeneity.

The ten different features for describing the binary regions that we have considered are: Convex Area, Eccentricity, Perimeter, Equivalent Diameter, Extent, Filled Area, Minor Axis Length, Major Axis Length, *R* and Solidity. Next, we give a brief description of each one.

- **Convex Area:** It is computed as the number of pixels of the smallest convex polygon that contains the region. The coordinates of such polygon (i.e., $(x_1, y_1), (x_2, y_2), (x_3, y_3), ..., (x_n, y_n)$) are arranged in the determinant form shown below. The coordinates are taken in counterclockwise order around the polygon, beginning and ending at the same point.

$$Area = \frac{1}{2} \begin{vmatrix} x_1 & y_1 \\ x_2 & y_2 \\ ... & ... \\ x_n & y_n \\ x_1 & y_1 \end{vmatrix} = \frac{1}{2}[(x_1y_2 + x_2y_3 + x_3y_4 + ... + x_ny_1) - (y_1x_2 + y_2x_3 + y_3x_4 + ... + y_nx_1)] \quad (1)$$

- **Eccentricity:** It is a scalar that specifies the eccentricity of the ellipse that has the same second central moments as the region (see Figure 3). The eccentricity is the ratio of the distance between the foci of the ellipse and its major axis length. Its value is between 0 and 1 (i.e., 0 and 1 are degenerate cases: an ellipse whose eccentricity is 0 is actually a circle, while an ellipse whose eccentricity is 1 is a line segment).

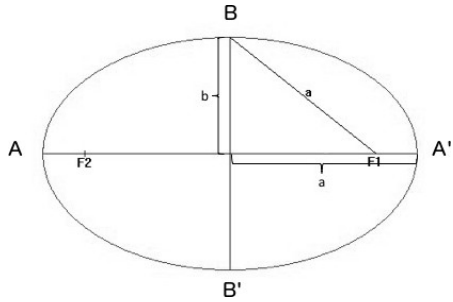

Figure 3. Ellipse with centre *O*. The foci coordinates are $F_2(-c, 0)$ and $F_1(0, c)$

The eccentricity of the ellipse is calculated as:

$$\sqrt{1 - \frac{b^2}{a^2}} \quad (2)$$

- **Perimeter:** This is the number of points in the contour.
- **Equivalent diameter:** Scalar that specifies the diameter of a circle with the same area as the region. It is computed as $\sqrt{\frac{4A}{\pi}}$, where $A$ is the area of the region.
- **Extent:** This feature specifies the ratio of the pixels of the region to the pixels in the bounding box around the region.
- **Filled Area:** It refers to the number of pixels belonging to the region after filling its possible holes.
- **Minor Axis Length:** Length of the segment $\overline{BB^t}$ (see Figure 3 (2b)) of the ellipse that has the same normalized second central moments as the region.
- **Major Axis Length:** Length of the segment $\overline{AA^t}$ (see Figure 3 (2a)) of the ellipse that has the same normalized second central moments as the region.
- $R = \frac{2b}{2a}$, where $2b$ and $2a$ stands for the lengths of the minor and the major axis, respectively, of the ellipse that has the same normalized second central moments as the region.
- **Solidity:** This feature indicates the proportion of the pixels in the smallest convex polygon that can contain the region, that are also in the region. Thus, this scalar is computed as $\frac{Area}{ConvexArea}$.

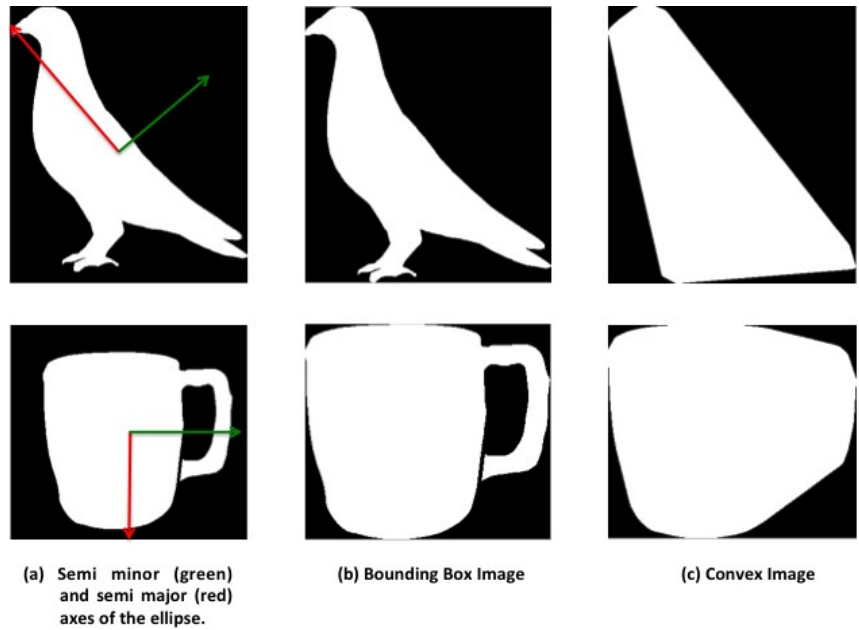

Figure 4. Two examples of the ShapeFeat features extraction. With the information shown in the first column, we can extract the eccentricity and R. With the information shown in (b), we can extract the extent. Finally, the smallest convex polygon shown in the third column (c) is necessary to extract the Convex Area and the Solidity.

Figure 4 depicts some of the relevant information used in this description method. In (a), the semi-minor and semi-major axes of the ellipse that has the same second central moments as the region are shown. This information is also taken into account to extract the eccentricity and $R$. In (b) the bounding box image crop is depicted, which is used to get the Extent property just calculating the ratio between the number of pixels in that image and the number of white pixels (region pixels). In (c), the smallest convex polygon is shown. That polygon is necessary to extract the Convex area value and the Solidity, which is the ratio between the number of white pixels in (b) and the number of white pixels in (c).

## 2.2 Contour Descriptor: B-ORCHIZ

B-ORCHIZ descriptor (García-Ordás et al. 2016) is based on the one developed by Anuar et al., called Zernike Moment Edge Gradient technique (ZMEG) (Anuar, Setchi, and kun Lai 2013).

It combines a global and a local descriptor taking advantage of both. A local descriptor describes a small area of the image around a point of interest: multiple local descriptors are used to match an image and make this process more robust for the comparison to be made. On the other hand, a global descriptor describes the characteristics of the whole image. Global descriptors are not as robust as local ones because any change in a small part of the image may cause the retrieval method to fail but it gives global information about the image. Combining them, i.e., global and local descriptors, the advantages of both strategies are kept.

Unlike ZMEG, B-ORCHIZ achieves invariance using Zernike moments' module up to the tenth order (i.e., 36 values) as global descriptor and the authors also proposed the Invariant Boundary Descriptor (IBD), composed by Invariant Edge Gradient Co-occurrence Matrix (IEGCM) and Boundary Orientations Chain (BOC), as local ones. The proposal is applied using the images resized to 128x128 pixels and interpolating the binary images between 0 and 255 allowing Zernike moments to be applied on a wider range of integer values. We address the reader interested in further information to (García-Ordás et al. 2016).

## 3. Combination of Shape and Contour descriptors

Humans usually take into account different combinations of image features to interpret them or to identify people, animals or objects. For example, the yellow color can be very useful to identify a banana in a image, but if shape properties are not taken into account too, it can be mistaken with a lemon or with the sun. This basic idea can be used in computer vision to increase the chance of success in image classification problems. In our work, we explore the combination of shape and contour features to improve our tool wear monitoring system. There are three well-known ways to merge descriptors: early fusion, intermediate fusion and late fusion (see Figure 5). Next, we describe with more detail each of them.

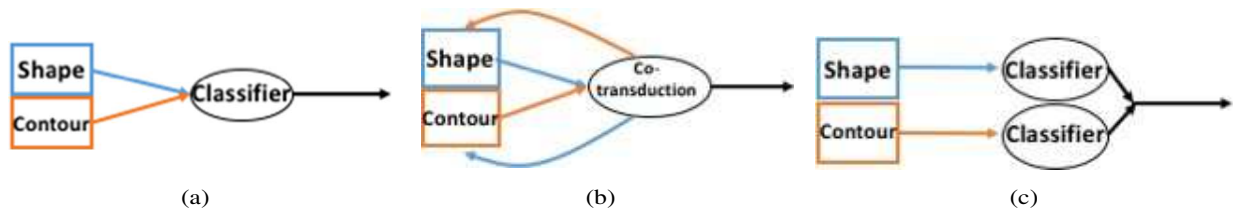

(a)        (b)        (c)

Figure 5. (a) Early fusion of the contour and shape descriptors. (b) Intermediate fusion using co-transduction method. (c) Late fusion carried out by means of Bayes Average.

### 3.1 Early fusion

Early fusion combines the features obtained using different techniques and produces a single feature vector to be used for the classification stage. The process inherently increases the size of the feature vector but it is one of the fastest fusion methods. In our case, we have used the simple concatenation of feature vectors. Thus, our input descriptor will be composed of all the B-ORCHIZ and ShapeFeat features.

## 3.2 Intermediate fusion: Co-transduction method

The Co-transduction method was originally proposed by Xiang Bai et al. (Bai et al. 2012) as a retrieval system. Their goal was to develop an algorithm to fuse different similarity measures for robust shape retrieval through a semi-supervised learning framework. The method was named Co-transduction, which was inspired by the Co-training algorithm (Blum and Mitchell 1998). In our case, the method has been adapted to deal with classification instead of retrieval using a k-Nearest Neighbors approach with the top-k retrieved images. Furthermore, we have combined two different descriptors instead of two different similarity measures to take advantage of all the information extracted by ShapeFeat and B-ORCHIZ. Given a query shape, the algorithm iteratively retrieves the most similar shapes using one description similarity matrix and assigns them to a pool for the other one to do a re-ranking, and vice versa.

Basically, the Co-transduction algorithm works as we can see in Algorithm 1. The reader interested in more details is addressed to (Bai et al. 2012).

---

**input :** a query object $x_1$, the database objects $X = x_2, ..., x_n$.
**output:** p nearest neighbors

**1** *Process:*;
**2** Create( $n \times n$ *probabilistic transition matrix $P_1$ based on one type of contour similarity (B-ORCHIZ)*);
**3** Create( $n \times n$ *probabilistic transition matrix $P_2$ based on one type of shape similarity (ShapeFeat)*);
**4** $Y_1 \leftarrow x_1$;
**5** $Y_2 \leftarrow x_1$;
**6** Create( *two sets $X_1, X_2$ such that $X_1 = X_2 = X$*)
**7** ; **for** $j \leftarrow 1$ **to** $m$ **do**
**8** | Use( $P_1$ *to learn a new similarity $sim^j_1$ by graph transduction when $Y_1$ is used as the query objects // ($j = 1, ..., m$ is the iteration index)*);
**9** | Add(*p nearest neighbors **from** $X_1$ **to** $Y_1$ based on the similarity $sim^j_2$ to $Y_2$*);
**10** | Add(*p nearest neighbors **from** $X_2$ **to** $Y_2$ based on the similarity $sim^j_1$ to $Y_1$*)
**11** | $X_1 \leftarrow X_1 - Y_1$
**12** | $X_2 \leftarrow X_2 - Y_2$ // (Then, $X_1$ and $X_2$ will be unlabeled data for graph transduction in the next iteration).
**13 end**

**Algorithm 1:** Co-transduction algorithm

---

## 3.3 Late fusion

The late fusion approaches use multiple classifiers to determine the output instead of just one as early fusion does. These methods try to combine the prediction score, i.e., the confidence of classifying the sample as positive, of all classifiers. Although it is very simple, this method has proved to be effective in improving performance of each individual classifier.

In this paper we have used the simple Bayes average (Ruta and Gabrys 2000; Bostrom 2007) as a method for obtaining a class probability distribution from the fused classifiers trained using contour (B-ORCHIZ) and shape (ShapeFeat) descriptors:

$$P_{LATEFUSION}(x \in C_i | x) = \frac{P_{SHAPEFEAT}(x \in C_i | x) + P_{BORCHIZ}(x \in C_i | x)}{2}, \quad (3)$$

where $x$ is one image and $C_i$ stands for one of all the possible classes it may belong to. $P_{SHAPEFEAT}$ and $P_{BORCHIZ}$ are the probability distributions for the different classifications carried out.

This method takes into account the information provided by all the classifiers and determine a decision based on the average of all of them, which avoid problems derived from a possible not fair classification.

## 4. Experimental results

The aim of this work is to characterize the insert state based on the wear region shape and explore whether or not adding more information, such as contour information, improves the categorization performance.

### 4.1 Dataset

In order to assess experimentally the proposed method, we have created an Insert dataset processing the images of 53 tools. The images were captured in a laboratory using a monochrome camera, model Genie M1280 1/3" with a 25mm optic AZURE and manual focus and aperture. The sensor has a resolution of 1280x960 pixels.

In order to improve the contrast in grey level images, we have used two lighting bars of red LEDs (BDBL-R82/16H). Images have been taken on the inserts disassembled from the cutting head and placed on a uniform background. Fig. 6 shows an example of the segmented inserts acquired under these conditions.

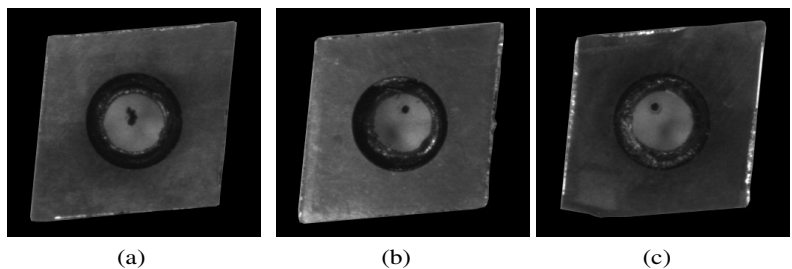

(a)          (b)          (c)

Figure 6. Examples of segmented inserts acquired under these experimental conditions. (a) Insert with low wear, (b) Insert with medium wear and (c) Insert that presents high wear.

Since the goal is to assess the wear level of each worn region, two preprocessing steps were carried out: the cropping of cutting edges and the wear region extraction.

From each insert, four cutting edge images have been extracted. This process starts by removing first the central portion of the insert after locating a circle whose center is in the centroid of the insert region and whose radius is $R = D/4.92$, where $D$ is the length of the major diagonal of the insert. The value 4.92 has been set empirically, based on the geometry of the insert. Thereafter, the cutting edges are detected using border detection filters and mathematical morphology operations, and extracted by cropping the subregion where they were located. Finally, the extracted cutting edges are rotated to place them in a horizontal position, as shown in Figure 7.

Subsequently, the wear region is extracted. Some examples of segmented wear regions are depicted in Figure 8.

Since the wear in the adjacent edges (i.e., the edges that do not appear in the horizontal axes in Fig. 7) may have an impact on the global wear of the insert, we have distinguished the wear in the two types of edges, so we have divided the dataset of wear regions into two: the "Complete edges" and the "Incomplete edges" (Insert-C and Insert-I, respectively). An example both kind of images is shown in Figure 9. The reader interested in more details is addressed to (García-Ordás et al. 2016).

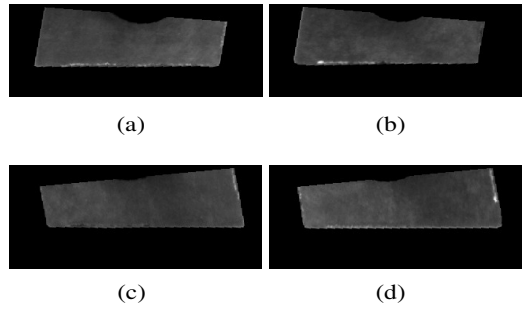

Figure 7. Examples of the North (a), South (b), East (c) and West (d) cutting edges cropped from the insert shown in Fig. 6(a).

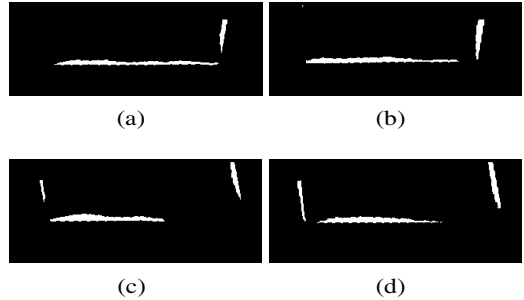

Figure 8. Examples of segmented wear regions of the North (a), South (b), East (c) and West (d) edges shown in Fig. 6(a)

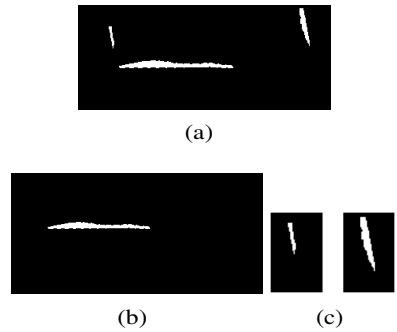

Figure 9. Examples of the types of inserts of a segmented image (a). In (b), the complete edge is depicted, whereas in (c) the incomplete edges are shown.

We followed a supervised learning approach to train the categorization model. An expert labelled the wear regions included in the dataset to generate the ground truth, carrying out two types of ratings depending on how many wear classes were taken into account. In the first one, three classes were assigned – Low (L), Medium (M) and High (H) wear – and two classes in the second one – Low (L) and High (H) wear –. In the three-class problem there were 126 edges with low wear, 260 with medium wear and 187 with high wear, whereas in the two-class problem, there were 260 edges with Low wear and 313 with high wear. The expert carried out the labelling process by means of a visual assessment, relying on his previous knowledge and experience. The decision of the expert depends on some considerations, like the shape, size and the distribution of the wear area, its location or how deep it is.

This labelling was carried out according to the three different stages identified in the well-known tool wear curve vs. time. In the three-class scenario (see Fig. 10(a)), the low class (L) corresponds to the rapid initial wear that occurs within the first machining minutes (break-in period). This is followed by a period with a uniform wear rate in a controlled way (steady state wear region) that was identified as medium (M) wear. Finally, the high wear label (H) refers to the stage with high wear that grows exponentially, in an uncontrolled way and with high risk of tool fracture. When we are dealing with only two wear levels, the short break-in period and the steady state wear region are included together in the low (L) wear class while the accelerated wear state where the tool is damaged is assigned to the high (H) wear class (Fig. 10(b)).

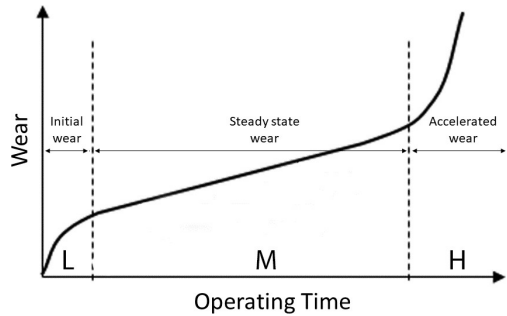

(a)

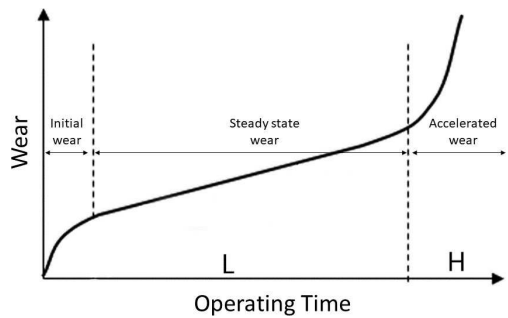

(b)

Figure 10. Tool wear vs. operating time. Correspondence between wear levels and the wear categories considered in this study: (a) Low / Medium / High categories and (b) Low / High Wear categories.

## *4.2 Experimental setup*

As we commented previously, to determine the wear of the complete edge by itself and also how it is influenced by the incomplete edges, we divided the whole Insert dataset into two: (a) Insert-C subset with the complete edges that correspond with the horizontal wear in the cropped images, and (b) Insert-I subset with the incomplete ones which orientations are vertical in the images.

To carry out the classification for these datasets with all the descriptors evaluated, we used a Support Vector Machine (SVM) classifier with intersection kernel. The validation has been carried out by means of stratified Monte Carlo cross-validation, which randomly splits the dataset into training and test data, both of them following the same distribution as the original dataset. For each split, the model is fit to the training data and the predictive accuracy is assessed using the test subset, that is, with images not seen during the calibration phase. The advantage of this method over k-fold cross validation is that the proportion of the training/test split is not dependent on the number of iterations. In this work, we have extracted 75% of the data for the training subset and the remaining 25% for the test subset. This process has been repeated 20 times and the final result is the accuracy averaged over these 20 runs.

Results are shown in terms of the accuracy achieved in the classification, which is the number of correctly classified cutting tools over the total number of cutting tools seen by the system, that, for the binary case is computed as shown in Equation (4).

$$accuracy = \frac{TP + TN}{TP + FP + FN + TN}, \quad (4)$$

where *TP* (true positives) represents the number of samples of the high wear class classified as high wear; *TN* (true negatives), number of samples of the low wear class classified as low wear; *FP* (false positives), samples classified as high wear when they belong to the low wear class and *FN* (false negatives) are samples classified as low wear when they actually show high wear.

We combined the two best descriptors (B-ORCHIZ and ShapeFeat) using early, intermediate (i.e., co-transduction) and late fusion techniques. The results were compared among each other and also with other classical descriptors. Co-transduction classifier uses k-NN. We assessed values of *k* equal to 3, 7, 9 and 11 obtaining the best results for *k* = 3 in all cases.

## 4.3 Results

In this section, we present the results obtained in all of our experiments divided in two main categories: Results achieved by each descriptor on its own (i.e., without combining them) and results combining both of them using different fusion techniques: early fusion, intermediate fusion and late fusion.

### 4.3.1 Results without fusion

The first step consists of assessing the descriptor based on shape presented in this paper (ShapeFeat) and other proposals based on contour features: B-ORCHIZ, ZMEG and aZIBO. Table 1 shows the performance in terms of accuracy.

Table 1. Classification accuracy (in %) of ZMEG, aZIBO, B-ORCHIZ and ShapeFeat using SVM with Intersection kernel for the complete, Insert-C, Insert-I dataset (from left to right) for two and three wear levels.

|  | Complete | | Insert-C | | Insert-I | |
|---|---|---|---|---|---|---|
|  | L-H | L-M-H | L-H | L-M-H | L-H | L-M-H |
| ZMEG | 83.74 | 75.87 | 85.58 | 76.44 | 83.52 | 79.40 |
| aZIBO | 84.44 | 78.85 | 87.02 | 76.92 | 84.89 | 82.14 |
| B-ORCHIZ | 87.06 | 80.24 | 87.02 | 81.25 | **88.46** | 82.69 |
| ShapeFeat | **88.70** | **80.67** | **93.37** | **81.35** | 88.41 | **84.12** |

It is clear that B-ORCHIZ achieves in all the datasets and all the classifications better results than ZMEG and aZIBO descriptors. The improvement for the complete dataset was more than a 3% in the binary classification and almost a 2% in the ternary one with respect the best descriptor of the state of the art (aZIBO). However, these results were outperformed by our proposed shape descriptor (ShapeFeat) in almost all the experiments. As it is shown in Table 1, the highest improvement was achieved with the Insert-C dataset, with an increment of more than a 7.29% of accuracy. In all the other cases, its behavior was very similar in comparison with the contour descriptors.

Feature ranking techniques play an important role to gain knowledge of data and identify the most relevant features. The ShapeFeat descriptor proposed in this paper includes interpretable shape features whose discriminant power has been analyzed. In order to identify the most relevant features, we performed a feature-ranking analysis following the well-known wrapper approach (Bolón-Canedo, Sánchez-Maroño, and Alonso-Betanzos 2013; Guyon et al. 2006). Wrapper methods use the performance of a learning algorithm (in this analysis, a Support Vector Machine (SVM) classifier) to assess the usefulness of a feature set. It iteratively discards features with the least discriminant power according to the classifier performance. Model performance was estimated by the Area under the ROC (Receiver Operating Characteristic) curve, where the ROC curve plots the true positive rate against the false positive rate. The ranking procedure was run on the Complete dataset and the following top-3 most relevant features were found: $R$ (i.e., the ratio between the minor and major axis), the major axis length and the extent.

### 4.3.2 Results with fusion of contour and shape descriptors

B-ORCHIZ, i.e., contour descriptor, and the proposed ShapeFeat, i.e., shape descriptor, are, to the best of our knowledge, the methods that achieve the best performance in the application of computer vision for tool wear monitoring so far. The big difference in the way both methods are built is very interesting because it makes possible to explore different fusion techniques to combine both of them.

In this section, we show the results of the three different fusion techniques applied over these two methods: early, intermediate and late fusion.

Early fusion was performed by concatenating B-ORCHIZ and ShapeFeat to create a new feature vector used as input for the SVM classifier. Intermediate fusion was carried out using the method Co-Transduction explained in Section 3.2. Late fusion was implemented combining the scores of the SVM classifier for B-ORCHIZ and ShapeFeat in order to determine the final response of the learner.

In Figure 11, results for each fusion method are shown. In order to compare the results with the original methods, we have also included B-ORCHIZ and ShapeFeat. Early fusion method achieves a very similar result to the B-ORCHIZ method on its own. This can be explained due to the high difference in size between both descriptors, which makes B-ORCHIZ to have more *weight* than ShapeFeat. Co-Transduction and Late fusion are invariant to the number of features because the fusion is carried out after the classification step has taken place. In almost all the experiments late fusion achieves a higher performance than the rest of the description techniques used on their own. However, in the binary classification using the Insert-C dataset, the high difference in accuracy between shape and contour descriptors conditions the result of the early and late fusion methods. The good performance of the Co-transduction method for this experiment is obtained because of the fusion algorithm. Whereas late fusion and early fusion are averagely influenced by both performances, in the Co-transduction method each descriptor is improved by the other. For this reason, although B-ORCHIZ does not show good results, it also improves the performance of ShapeFeat instead of decreasing it in the fusion step.

### 4.3.3 Descriptor fusion vs baseline descriptors

We also compare our proposal with other classical descriptors like Bag of contour fragments (BCF) (Wang et al. 2014b), Histogram of oriented gradients (HOG) (Dalal and Triggs 2005) and Shape Context (SC) (Belongie, J.Malik, and Puzicha 2002). As it is shown in Table 2, our combined method outperforms the state of the art methods in all cases, achieving improvements of more than 82%, 35% and 31% with respect to SC, HOG and BCF methods, respectively, in all the cases evaluating the L-M-H classification performance for the complete dataset.

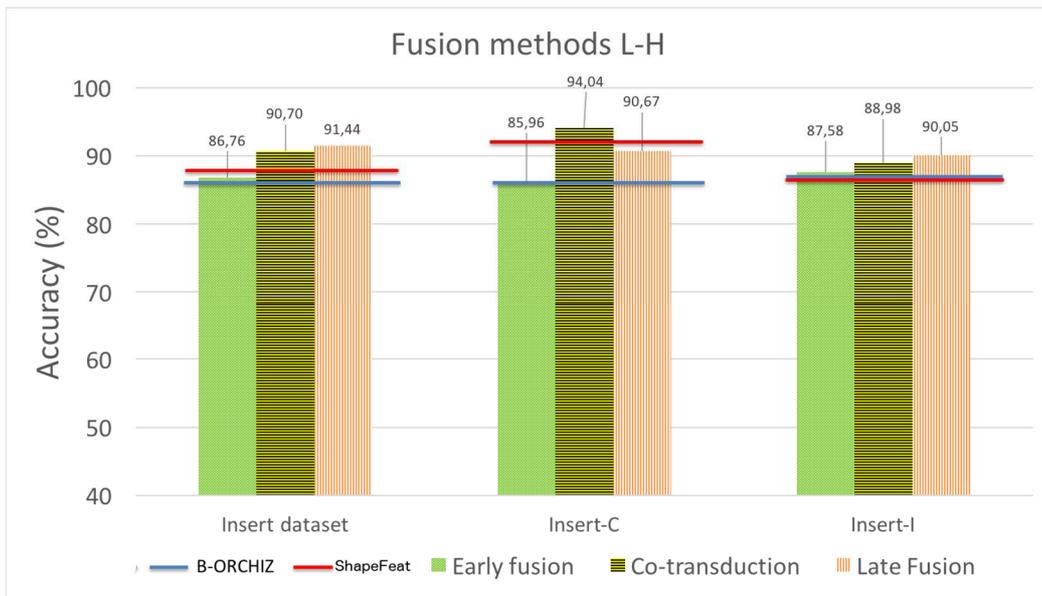

(a)

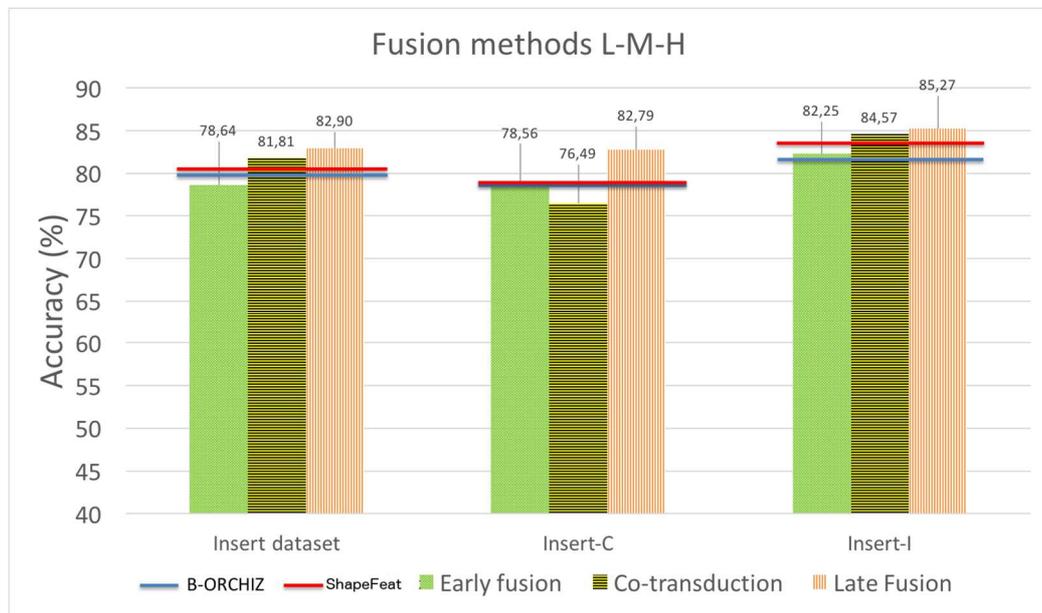

(b)

Figure 11. Results for binary (a) and ternary (b) classification using fusion methodology with B-ORCHIZ and ShapeFeat. Furthermore, results achieved by the descriptors on their own are represented as straight lines over the graphs.

It is noteworthy that the tool assessment we propose takes place when the head tool is in the parking position, between the machining of two consecutive plates, ensuring that the tool state is suitable for the next operation because it is not in the High wear category. The resting time of milling head tools lies between 5 and 20 minutes, what makes our approach appropriate to be implemented in this real production environment.

Table 2. Classification accuracy in % of the combination of ShapeFeat with BORCHIZ and other descriptors like Bag of contour fragments (BCF), Histogram of oriented gradients (HOG) and Shape Context (SC).

|  | Complete | | Insert-C | | Insert-I | |
|---|---|---|---|---|---|---|
|  | L-H | L-M-H | L-H | L-M-H | L-H | L-M-H |
| Late fusion | **91.44** | **82.90** | **90.67** | **82.79** | **90.05** | **85.27** |
| BCF | 76.76 | 63.11 | 77.88 | 49.52 | 80.77 | 74.45 |
| HOG | 76.80 | 60.35 | 80.19 | 51.06 | 76.65 | 66.15 |
| SC | 54.58 | 45.45 | 68.75 | 41.83 | 69.51 | 48.08 |

## 5. Conclusion

In this work we present a new approach to conduct an automatic assessment of the wear level in milling inserts. Evaluating the most recent studies connected with this field following a computer vision based approach, the best technique developed so far for evaluating the status of the inserts is the B-ORCHIZ descriptor. This method is based on the description of the contour. There are, however, many potential features that can be used to represent an image. In order to evaluate other possibilities, in this work we proposed ShapeFeat, ten features that describe the shape of the binary images. Experimental results showed that this descriptor based on shape outperformed previous proposals. Additionally, a feature-ranking analysis with a wrapper approach revealed that the most relevant shape features are: R (the ratio between the minor and major axis), the major axis length and the extent.

Taking into account the good performance of contour and shape descriptors and, bearing in mind that humans combine different characteristics for image recognition tasks, we explored the combination of these two approaches. The combination was carried out using three fusion methods: early, intermediate (which combines the similarity matrices of each method to improve the classifier of the other one) and late fusion. The accuracy achieved using late fusion of both descriptors outperformed individual performance: accuracy was 91.44% for binary classification using the whole dataset and 82.90% in the low-medium-high one. These empirical results provide evidence that this approach is a very promising opportunity for developing an automatic wear monitoring system in edge profile milling processes, saving time in the insert review process, its associated costs and avoiding the possible errors due to the subjectivity of the human evaluation.

The proposed methodology has been assessed on a specific type of inserts, whereas a huge variety of situations may be found in real industrial environments, i.e., inserts with different shapes or different materials being machined. Despite of being assessed for specific inserts, our proposal can be extended to any other tool or material. For example, transferring this proposal to other production environment with other materials being machined, requires the collection of a representative tool image dataset labelled by experts in terms of the wear degree, for the subsequent training of the categorization module. When the tool changes, the image preprocessing should be adapted. The features proposed in this paper can thereafter be extracted from the wear region and used to create the model that automatically classifies the wear level of new unseen tools.

Addressing the problem of automatically detecting the type of tool wear by means of a similar framework based on computer vision is part of our future research.

## Acknowledgements

This work has been supported by the research project with ref. DPI2012-36166 from the Spanish Ministry of Economy and Competitiveness and the PIRTU program of the Regional Government of Castilla y León.